\documentclass[sigconf,screen,nonacm]{acmart}
\AtBeginDocument{%
  \providecommand\BibTeX{{%
    \normalfont B\kern-0.5em{\scshape i\kern-0.25em b}\kern-0.8em\TeX}}}

\usepackage[linesnumbered,boxed]{algorithm2e}

\begin{document}

\title{Ranking labs-of-origin for genetically engineered DNA using Metric Learning}

\author{Fernando H. F. Camargo}
\authornote{Both authors contributed equally to this research.}
\email{fernando@amalgam.ai}
\affiliation{%
  \institution{Amalgam}
  \streetaddress{P.O. Box 1212}
  \city{Austin}
  \state{Texas}
  \country{USA}
  \postcode{43017-6221}
}
\author{Igor Muniz Soares}
\email{igor@amalgam.ai}
\authornotemark[1]
\affiliation{%
  \institution{Amalgam}
  \streetaddress{P.O. Box 1212}
  \city{Austin}
  \state{Texas}
  \country{USA}
  \postcode{43017-6221}
}

\author{Adriano Marques}
\email{adriano@amalgam.ai}
\affiliation{%
  \institution{Amalgam}
  \streetaddress{1 Th{\o}rv{\"a}ld Circle}
  \city{Austin}
  \state{Texas}
  \country{USA}}

\renewcommand{\shortauthors}{Camargo and Muniz, et al.}

\begin{abstract}
  With the constant advancements of genetic engineering, a common concern is to be able to identify the lab-of-origin of genetically engineered DNA sequences. For that reason, AltLabs has hosted the Genetic Engineering Attribution Challenge to gather many teams to propose new tools to solve this problem. Here we show our proposed method to rank the most likely labs-of-origin and generate embeddings for DNA sequences and labs. These embeddings can also perform various other tasks, like clustering both DNA sequences and labs and using them as features for Machine Learning models applied to solve other problems. This work demonstrates that our method outperforms the classic training method for this task while generating other helpful information.
\end{abstract}





\maketitle

\section{Introduction}\label{sec:intro}

Genetic engineering or genetic modification is one of the fastest-growing fields in biotechnology. It is responsible for manipulating an organism's genes. It is a well-known technology with applications in numerous fields, bringing several benefits to science, medicine, and industry, which is why it is receiving increasing investments. Changing DNA makes it possible to create drugs and vaccines while reducing costs and time to market in health and industrial applications. However, alongside the benefits, there are possible situations of improper usage. Modification of genes can lead to the accidental or intentional release of new and dangerous biological agents.

Detecting genetic modifications and attributing them to a specific origin is a recent and open problem. Tools have been developed for detecting genetic engineering through genetic analysis of the organism genome. This is the case of FELIX (Finding Engineering-Linked Indicators) \cite{felix}, a program developed by IARPA that can rapidly detect indicators of genome engineering and has recently demonstrated its power by demystifying the rumors that the COVID-19 virus was created in a laboratory. However, assigning the origin of a specific engineered DNA to a lab is a more complex challenge. 

AltLabs, a not-for-profit organization,  hosted the Genetic Engineering Attribution Challenge \cite{competition}, providing the participants with a dataset of 63,017 DNA sequences (with their phenotype characteristics) designed by 1,314 different labs. The goal was to design machine learning models to, given a DNA sequence, identify the most likely lab-of-origin. Since it is tough to predict the correct lab because they might use similar design techniques, they decided to evaluate the solutions using top ten accuracy \cite{Alley2020.08.22.262576}. It means that the model needs to place the correct lab-of-origin within the ten most likely labs, according to the scores. It turns this problem into a ranking problem instead of a simple classification, which might ask for different techniques.

During the competition, we decided to create two branches. The first is a traditional approach using classification models (with a softmax to output probabilities for each lab). The second is the use of ranking models through Metric Learning \cite{kulis2013} (more specifically, Triplet Networks \cite{hoffer2018deep}). This one has the goal of learning how to extract embeddings\footnote{Embedding is a vectorial representation of some entity composed of latent features. Humans cannot easily understand these features, but they are beneficial for machine learning models. When we train a model to extract embeddings, the goal is to place similar entities closer together in the latent space.} (also known as feature vectors) from DNA sequences and, at the same time, learn the embeddings of the labs. A similarity measure (cosine similarity) is then applied to generate the score between a DNA sequence and a lab. This score, instead of trying to mimic a probability, describes how similar they are.

Despite the difference in how we train those two kinds of models, we designed them very similarly. Both of them shared the same preprocessing steps, and the first layers were almost the same. They diverged a little bit because some techniques worked better for one than the other. Nevertheless, the overall structure of these models was mostly the same. Even though they were very similar, our triplet networks consistently outperformed our classification models, indicating they are more suitable for ranking tasks.

Furthermore, it is not only about raw accuracy. Suppose a new sequence from an unknown lab goes through a regular classification model. In that case, it will probably attribute it to one of the known labs while giving no clue about its uncertainty. It is a known problem and has encouraged research in an area called Out-Of-Distribution Detection \cite{chen2020}\cite{ren2019}. On the other hand, our triplet network is more robust to this problem. The embedding extracted from such a sequence might differ significantly from all the labs' embeddings, generating low similarity scores. This way, we could set a threshold and decide that it comes from an unknown lab if the score is below it. There is also an embedding learned for an unseen lab. We train the model to push it away from all the known labs. So, suppose the new sequence's embedding is more similar to this one. In that case, we can also assume an out-of-distribution sequence. In summary, our proposed solution has a high accuracy while also dealing with new labs and providing embeddings for clustering and other tasks.

\begin{figure}[h]
  \centering
  \includegraphics[width=0.5\textwidth]{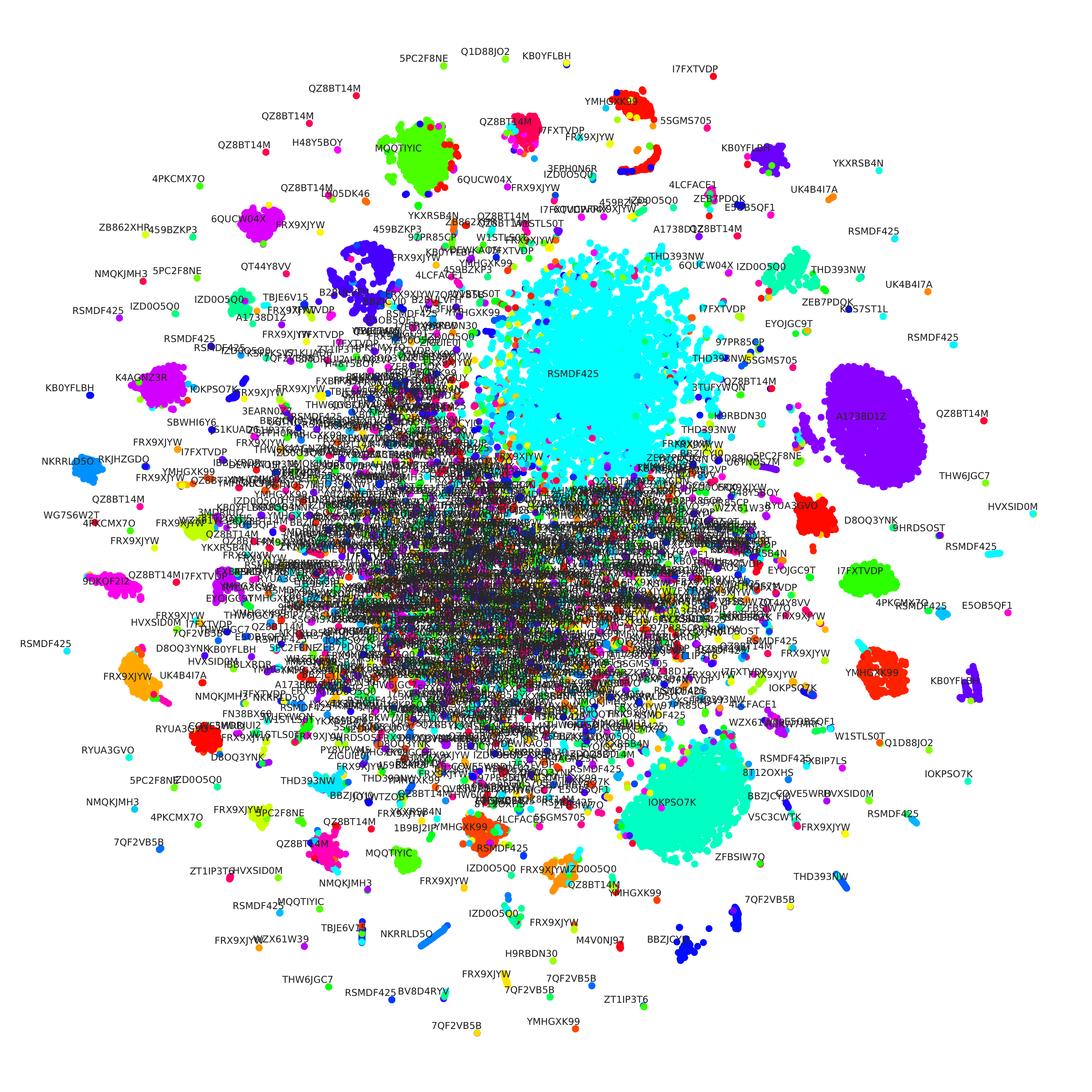}
  \caption{We used t-SNE \cite{hinton2002} to project 200D into 2D and produce this Figure. Each circle represents a DNA sequence, and the color shows the lab-of-origin of such sequence. The lab is also presenting using a text with its ID. We can also see that there are many sequences very similar in the middle, even though they come from different labs. We theorize that they are from small labs that use similar design techniques. It is also worth noting that dimensionality reduction causes some losses, which might also explain it.}
  %\Description{TODO}
  \label{fig:tsne}
\end{figure}
\section{Proposed Method}

\subsection{Implementation Details}

Our proposed method consists of different preprocessing techniques proved to be efficient for DNA sequences in our experiments, along with a Neural Network. As a preprocessing step, we used Byte Pair Encoding (BPE) algorithm \cite{gage1994} to compress the DNA sequences. It works by looking for common patterns and unifying them into tokens, increasing the vocabulary while reducing the sequences' size. It turned our vocabulary of 4 DNA bases into 1001 different tokens.

To deal with the fact that DNA sequences are circular, we used a circular shift data augmentation. This augmentation was used during training and inference, doing what is known as Test Time Augmentation. It means we feed the model with multiple versions of the same sequence (randomly circular shifted). We then take the average of the outputs. Further, to decrease the model's complexity, we also limited the sequence to 1000 tokens. This operation runs after the circular shift, so the model receives different 1000 tokens sequences per execution and combines the results.

\begin{figure*}[t!]
  \centering
\includegraphics[width=\linewidth]{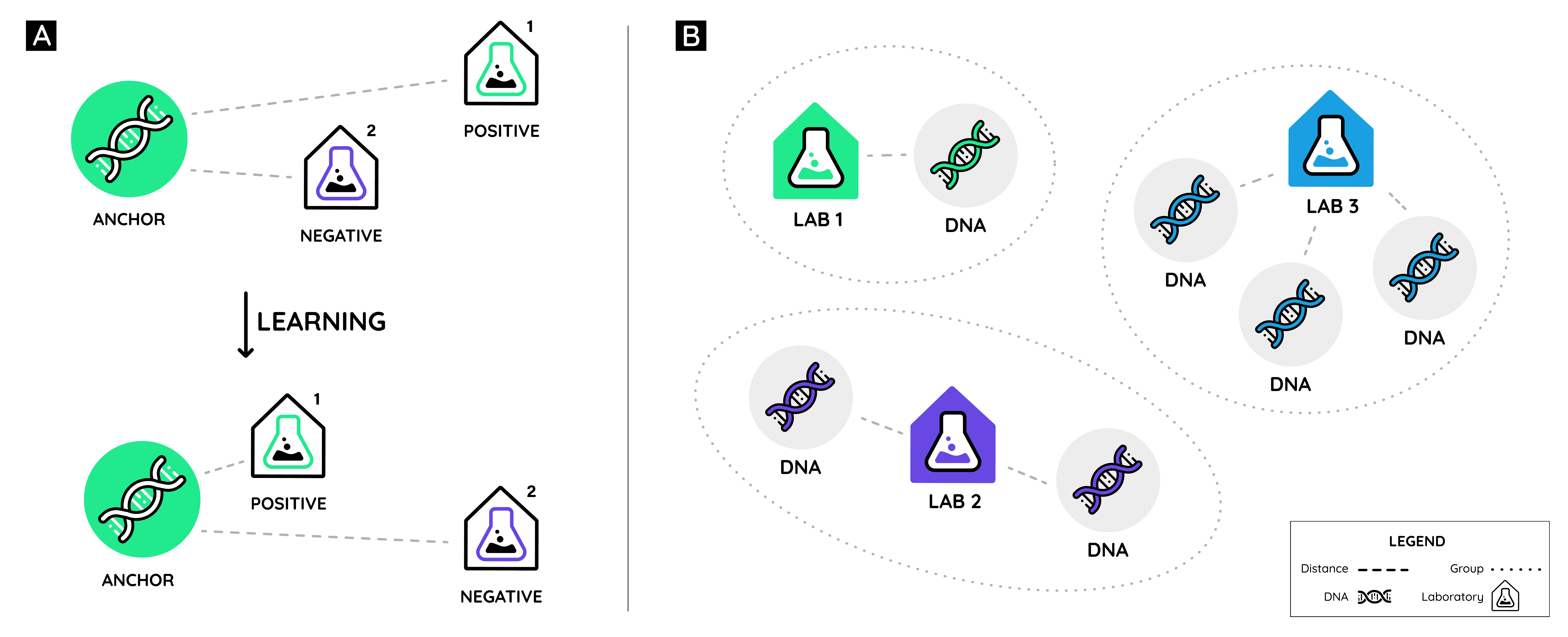}
  \caption{(A) In the beginning, the negative (another lab) might be closer to the anchor (DNA) than the positive (lab-of-origin). During training, we pull the anchor and lab-of-origin towards each other while pushing the negative away. This way, the positive will be closer to the anchor, and the negative will be farther. (B) In the end, labs and their DNA sequences will be closer to each other, forming groups. We can also expect both labs and DNA sequences to be closer to other similar ones.}
%   \Description{Diagram flow of \textit{MARS-Gym}, from dataset ingestion for environment generation to the MDP simulation.}
  \label{fig:triplet}
\end{figure*}

Both types of models are composed of a Convolutional Neural Network with multiple kernels of different sizes, as proposed by \citeauthor{kim2014}. We use it to extract features from the sequence and concatenate them with the binary features provided in the dataset. The difference between the classification and triplet network models are after this part.

The classification model takes these features and pass them through a dense layer and then the output layer. This output layer is a dense layer with softmax as the activation function, resulting in probabilities for each lab. On the other hand, the triplet network passes these features through a dense layer that will generate our sequence embedding. In parallel, we have an embedding layer that will learn the lab embeddings. For last, we use a cosine similarity measure between the embeddings to give us the final output.

While we train the classification model using regular Supervised Learning, we train the Triplet Network differently. We create triplets of anchors, positives, and negatives and use them for training the model. In a Face Recognition task, the anchor would be one picture of one person, the positive would be another picture of the same person, and the negative would be any picture of another person. The model learns to pull the anchor and positive together (making pictures of the same person produce similar embeddings) while pushing the negative away. In our case, the DNA sequence is the anchor, while the positive is the source lab, and the negative is another lab. Our goal is to generate embeddings in which the DNA sequences are close to their source lab and far from other labs and their sequences. Figure \ref{fig:triplet} summarizes this process.

To generate these triplets, we use the dataset to provide us with the anchor (sequence) and positive (source lab). We then use a technique known as Hard Negative Mining \cite{hermans2017} to select the negative (another lab). It means that instead of selecting a random lab as a negative example, we select the most challenging example. It would be the lab that is currently closer to our sequence in the latent space.

\subsection{Triplet Network Advantages}

After training, we can use the Triplet Network to extract embeddings from DNA sequences. It also contains a table of embeddings for each lab present in the dataset and one embedding for an "unseen" lab. These embeddings can be helpful beyond the expected usage (ranking possible labs-of-origin given a DNA sequence). We can also use them to compare and cluster DNA sequences and labs in terms of design style, as shown in Figure \ref{fig:tsne}. Speaking of which, we applied K-Means \cite{lloyd1982} to cluster the lab embeddings to find out what is a good number of groups for the 1,314 labs available. Our result, using the Elbow Method, is that we can comfortably put them into nine groups. Such clustering capabilities of our model could be instrumental in investigating the relationship between these labs.

We also expect this model to be very robust for unseen labs. As pointed out in Section \ref{sec:intro}, we can set a threshold and use the unseen embedding to detect that a DNA sequence comes from an unknown lab. Besides that, suppose that an accidental release happened and we need to find out the responsible. Even though some labs are not part of our training dataset, we could ask for some samples of sequences designed and extract their embeddings. We could average these embeddings to generate an embedding for the lab. With this embedding at hand, we could compute this lab's similarity with the DNA sequence that was accidentally released. It is worth noting that we can do that without requiring the model to be retrained.

For last, these embeddings are also very feature-rich, which means we can use them as input for other machine learning models, tackling other problems. For example, we know that it is common to design DNA sequences derived from multiple different sources. Suppose we want to identify the composition of a given DNA. We could prepare a dataset and use our embeddings to serve as inputs for another model. It would drastically accelerate the creation of such a model.

\begin{algorithm}[h]
  \SetAlgoNoEnd
  \DontPrintSemicolon
  \KwIn{$lab\_indices$ (B,), $anchor\_embeddings$ (B, L), $positive\_embeddings$ (B, E)}
  \KwOut{Batch of $negatives$ of shape (B, E)}
  $ all\_lab_indices \leftarrow $ indices of all labs repeated by B // (B, L) \;
  $ negative\_labs\_mask \leftarrow $ matrix of boolean values with $True$ for every lab // (B, L) \;
  $ negative\_labs\_mask[:, lab\_indices] \leftarrow False$ // (B, L) \;
  // At this point, we have a mask with True for the negative labs and False otherwise \;
  $ all\_negative\_lab\_indices \leftarrow $ $all\_labs[negative\_labs\_mask]$ reshaped // (B, L-1) \;
  $ all\_negative\_labs\_embeddings \leftarrow lab\_embeddings[all\_negative\_labs]$ L2 normalized // (B, L-1, E) \;
  $ anchor\_similarities \leftarrow $ dot product between $anchor\_embeddings$ and $all\_negative\_labs\_embeddings$ // (B, L-1) \;
  $ hardest\_negative\_lab_indices \leftarrow $ argmax of $anchor\_similarities$ // (B,) \;
  $ negative\_embeddings \leftarrow all\_negative\_labs\_embeddings[:, hardest\_negative\_lab\_indices]$ // (B, E) \;
  \Return{$negative\_embeddings$}
  
  \caption{Algorithm for Hard Negative Mining using tensors. The shape of each tensor is at the end of each line as a comment, being B batch size, E the embedding dimension, and L is the number of labs. It is worth noting that we L2 normalize all the embeddings.}\label{alg:hard_negative_mining}
\end{algorithm}
\section{Results}

% triplet: 90.39\%
% softmax: 89.36\%
% final: 91.67\%
% position: 10th

During the competition, we tuned the classification model and triplet network to improve their performance. As a result, our best triplet network got 90.39\% top ten accuracy in the test dataset. In contrast, our best classification model got 89.36\%. Thus, it shows that the triplet network displays many advantages and performs better than a classification model in such a ranking task. It is worth noting that we are comparing both of them using a very similar architecture. At the end of the competition, other competitors created more accurate classifiers. However, we theorize that if we trained their architecture with our scheme, they would be even more precise.

For the competition, we decided to combine these models into an ensemble to boost our accuracy further. However, since they give very different outputs, this approach posed several challenges. We first tried to compute each lab's average rank position known as Borda voting rule \cite{borda1781}. However, the results were worse than the individual models. We then decided to try the Copeland voting rule \cite{saari1996}, which performed much better. Our final submission using this ensemble got 91.67\% top ten accuracy and the 10th position in the competition. We also got the 2nd position in the Innovation Track, meaning that the organizers recognized our solution as innovative.
\section{Conclusions and Future Work}

Metric Learning is quite common in other areas (like Face Recognition and Recommender Systems). However, to the best of our knowledge, it has never been used in this context. We showed an innovative solution with promising results, outperforming a classification model trained with Supervised Learning. Moreover, it can also generate embeddings as a by-product, which are very useful.

We believe that more advanced techniques could be applied to extract features from the DNA sequences to improve them even further. New techniques like Transformers and Graph Convolutional Networks would be good candidates for this task. With better feature extraction, the embeddings would have higher quality, improving overall accuracy.

% \begin{acks}
% \end{acks}

\bibliographystyle{ACM-Reference-Format}
\bibliography{bib/references.bib}

%\appendix

\end{document}